\documentclass[twoside,twocolumn]{article}

\usepackage{wscg}           
\RequirePackage{ifpdf}
\ifpdf
 \RequirePackage[pdftex]{graphicx}
 \RequirePackage[pdftex]{color}
\else
 \RequirePackage[dvips,draft]{graphicx}
 \RequirePackage[dvips]{color}
\fi
\usepackage{nopageno}       

\usepackage{censor} 
\usepackage{bm} 
\usepackage{physics} 
\usepackage{amssymb} 
\usepackage{mathtools}	
\usepackage{tabularx} 
\usepackage{algorithm} 
\usepackage[noend]{algpseudocode} 
\usepackage{multirow} 
\usepackage{tabularx,booktabs} 
\usepackage{todonotes}
\usepackage{graphicx}
\usepackage{amsmath}
\usepackage{amssymb}
\usepackage{booktabs}
\usepackage{bm}
\usepackage{todonotes}
\usepackage{tabularx}
\usepackage{xcolor}
\usepackage{acronym}
\usepackage{todonotes}
\usepackage{subcaption}
\usepackage{balance}

\usepackage{xcolor}

\newcolumntype{Y}{>{\raggedleft\arraybackslash}X}

\acrodef{ANE}[ANE]{Apple Neural Engine}
\acrodef{AR}[AR]{augmented reality}
\acrodef{BTF}[BTF]{bidrectional texture function}
\acrodef{CNN}[CNN]{convolutional neural network}
\acrodef{HDR}[HDR]{high-dynamic range}
\acrodef{IBL}[IBL]{image-based lighting}
\acrodef{LDR}[LDR]{low-dynamic range}
\acrodef{MLP}[MLP]{multilayer perceptron}

\usepackage[capitalize]{cleveref}
\crefname{section}{Sec.}{Secs.}
\Crefname{section}{Section}{Sections}
\Crefname{table}{Table}{Tables}
\crefname{table}{Tab.}{Tabs.}

\StopCensoring		

\title{Real-time Light Estimation and Neural Soft Shadows for AR Indoor Scenarios}

\author{
	\parbox{0.25\textwidth}{\centering
		\censor{Alexander Sommer$^{1}$}\\[1mm]
		\censor{alexander.sommer@hs-rm.de}
	}
	\hspace{0.05\textwidth}
	\parbox{0.25\textwidth}{\centering
		\censor{Ulrich Schwanecke$^{1}$}\\[1mm]
		\censor{ulrich.schwanecke@hs-rm.de}
	}
	\hspace{0.05\textwidth}
	\parbox{0.25\textwidth}{\centering
		\censor{Elmar Schoemer$^{2}$}\\[1mm]
		\censor{schoemer@uni-mainz.de}
	}
	\\[8mm]
	\parbox{\textwidth}{\centering 
		\censor{$^1$ Computer Vision and Mixed Reality Group, RheinMain University of Applied Sciences} \censor{Wiesbaden R\"usselsheim, Germany}} \\
		\censor{$^2$Institute of Computer Science, Johannes Gutenberg University Mainz, Germany}
		\vspace{5mm}
}

\usepackage{url}
\makeatletter
\def\Uslash{\mathbin{\mathchar`\/}\@ifnextchar{/}{\kern-.15em}{}}
\g@addto@macro\UrlSpecials{\do \/ {\Uslash}}
\def\Ucolon{\mathbin{\mathchar`:}\@ifnextchar{/}{\kern-.1em}{}}
\g@addto@macro\UrlSpecials{\do : {\Ucolon}}
\makeatother

\begin{document}

\twocolumn[{\csname @twocolumnfalse\endcsname

\maketitle  

\begin{abstract}
\noindent
We present a pipeline for realistic embedding of virtual objects into footage of indoor scenes with focus on real-time AR applications. Our pipeline consists of two main components: A light estimator and a neural soft shadow texture generator. Our light estimation is based on deep neural nets and determines the main light direction, light color, ambient color and an opacity parameter for the shadow texture. Our \emph{neural soft shadow} method encodes object-based realistic soft shadows as light direction dependent textures in a small MLP. We show that our pipeline can be used to integrate objects into AR scenes in a new level of realism in real-time. Our models are small enough to run on current mobile devices. We achieve runtimes of 9ms for light estimation and 5ms for neural shadows on an iPhone 11 Pro.

\vspace{0.5em}


\vspace*{1.0\baselineskip}

\end{abstract}
}]


\section{Introduction}
\label{sec:intro}

\copyrightspace{80-903100-7-9}{2005}{January 31 -- February 4}

We propose a method for realistically inserting virtual objects into indoor scenes in the context of augmented reality applications. Thereby we first estimate the current lighting situation in the scene from a single RGB image captured by the camera of, for example, a mobile device. Then we use this information to insert the virtual object into the existing scene as plausibly and realistically as possible.
\begin{figure}
	\centering
	\includegraphics[width=1.0\linewidth]{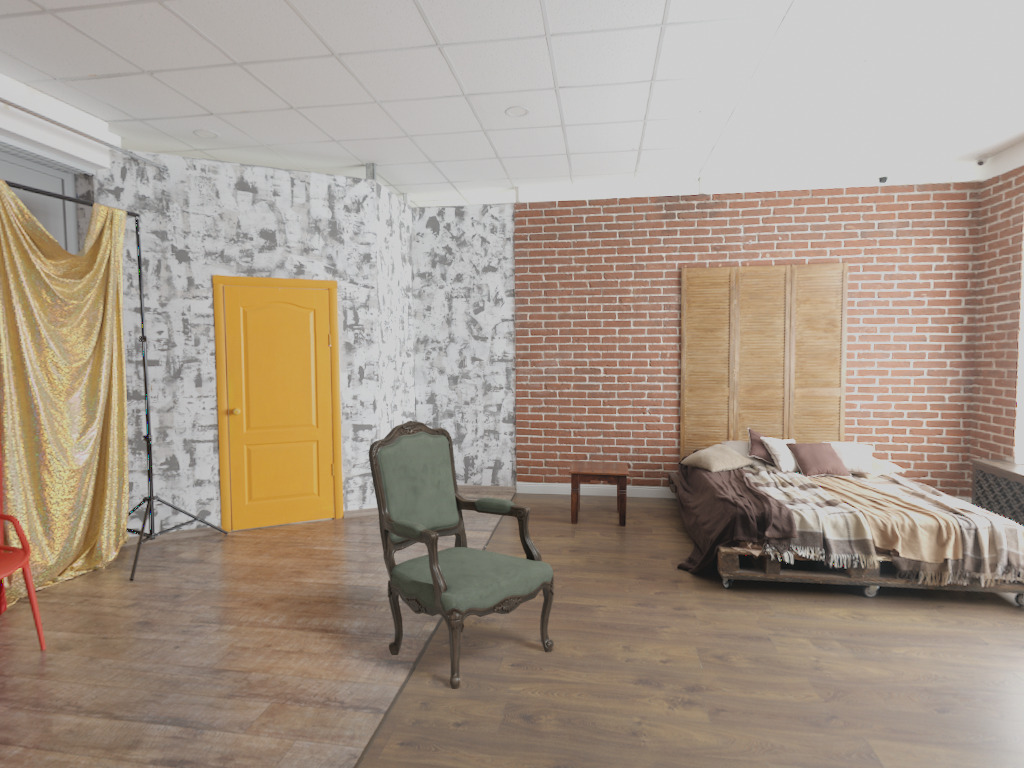}
	\caption{Example application of our pipeline: The light estimation determines the light direction and ambient color for rendering the inserted object. Based on the determined light direction, additional neural soft shadows are generated to create a realistic shadow cast as texture.}
	\label{fg:example_insert}
\end{figure}

The light situation in an existing scene can be caputured by placing a light probe at the position of the image. This can create a 360$^\circ$ \ac{HDR} panorama, also called environment map, of the scene. Such an \ac{HDR} image contains a large amount of information about bright and dark areas that would be clipped as black or white in an ordinary 8Bit \ac{LDR} image. Since the map contains information about the illumination of each direction of the scene at a given point, this environment map can be utilized to illuminate an object as if it were in the scene using methods from \ac{IBL}~\cite{Debevec98}. Some techniques have been developed to estimate this environment map from a single limited field-of-view \ac{LDR} image without additional 360$^\circ$ cameras using neural nets and deep learning~\cite{Gardner17, Song19, Somanath21}. However, such an environment map is only valid for a single specific point in the scene. Moreover \ac{IBL} techniques can be used to realistically illuminate objects with spatial varying light. Shadows in \ac{IBL} are created by tracing the path of light and its interaction with other objects in the scene. While this produces a very realistic shadow cast, a large number of path traces is required. This is computationally intensive and therefore not suitable for real-time applications on mobile devices.

Alternatively, parametric models exist that describe the light sources as physical objects in a 3D scene. In contrast to an environment map from \ac{IBL}, these models are valid for the entire scene. In the simplest case, such a parametric model can be for example a directional light with a fixed direction. Parametric light sources have a long history in computer graphics and there are several methods to efficiently calculate the shadow cast by objects. However, they often have to be modeled manually by a 3D artist for an existing scene. Recently, methods came up to estimate parametric lights directly from an input image using neural networks~\cite{Cheng18, Garon19, Gardner19}. Our work takes up on these methods. We restrict ourselves to reliably predict the main light direction at a given point from a limited field-of-view \ac{LDR} indoor RGB camera image and additionally determine the light color as well as the ambient color.

Especially in indoor scenes, the lighting situation is very complex. For the realistic overlay of virtual objects in the context of AR~\cite{Ismar14}, it makes a big difference whether a realistic or visual convincing shadow cast is present. Many other light estimation works map the existing lighting situation, but are only able to realistically insert virtual objects through offline rendering, e.g. ray tracing. Most shadows in indoor scenes are soft shadows as they are caused by light objects in a relatively short distance with a certain surface area. They are much more complex to compute than hard shadows caused by a quasi infinitely distant light source like the sun in outdoor scenes. We present a method to use the estimated light direction from the previous part to generate realistic indoor soft shadows in real-time. For this purpose we present a novel approach to encode precomputed ray-traced soft shadows using a neural network. This small network can be queried in real-time to generate a shadow texture depending on the light direction (see \cref{fg:example_insert}).

\noindent Our \emph{main contributions} are as follows:
\begin{enumerate}
	\item An improved deep neural network for parametric light direction estimation in indoor scenes.
	\item A new method for encoding shadow textures in an MLP that is memory friendly and fast to query.
	\item A complete pipeline for light estimation and shadow creation for real-time AR applications on mobile devices.
\end{enumerate} 
%
%

\section{Related Work}
\label{sec:relWork}

Existing work related to ours can be roughly divided into two categories. On the one hand, research in the area of estimating the existing light situation in the real scenery and, on the other hand, research on how to use this information for the realistic insertion of virtual objects into the augmented reality.

\textbf{Light estimation} is a classical problem in the field of computer vision or computer graphics as a subarea of 3D scene reconstruction. An accurate determination of the existing lighting conditions is crucial for a convincing insertion of virtual objects into the real environment.

Classic approaches usually require multiple images and/or more detailed knowledge about the underlying scene geometry. For example, Debevec and Malik showed how the omnidirectional \ac{HDR} radiance map can be reconstructed using multiple shots of a reflecting sphere with different exposure settings~\cite{Debevec97} and how to render synthetic objects into real scenes~\cite{Debevec98}. Lombardi and Nishino~\cite{Lombardi16} showed how illumination can be reconstructed from a single image of an object with known geometry. Balc\i\ and G\"ud\"ukbay~\cite{Balci17} reconstructed illumation based on the shadows in scenes that were mainly illuminated by the sun. Baron and Malik~\cite{Barron2013} reconstructed not only the illumination but also the geometry and reflectivity of an unknown object from an image using shape priors. Lopez-Moreno et al.~\cite{Lopez-Moreno13} presented an approach based on heuristics that does not require geometric knowledge.

With the rise of machine learning based approaches the need for information about the scenery could be further reduced. There exist quite some work that estimate lighting information and environment maps. For example, Hold-Geoffroy et al.~\cite{Hold-Geoffroy17} used a deep neural net to predict the illumination in outdoor scenes from a single image by relying on a physically-based sky model. Gardner et al.~\cite{Gardner17} estimated an \ac{HDR} illumination map for indoor scenes also from a single image by splitting the process into light position estimation and \ac{HDR} intensity estimation. Song and Funkhouser~\cite{Song19} used a multi-stage approach to predict a 360$^\circ$ \ac{LDR} map from a single image and completed geometry and intensity on \ac{HDR} scale. Somanath and Kurz~\cite{Somanath21} predicted a true \ac{HDR} map from a single camera image in a single stage approach tailored to mobile \ac{AR} real-time applications. Other approaches focus more on estimating light in form of low dimensional parameters. Garon et al.~\cite{Garon19} used spherical harmonic coefficients as light model. Cheng et al.~\cite{Cheng18} also used spherical harmonics for their light model, but used the images from the front and rear camera for the estimation. 

Gardner et al.~\cite{Gardner19} described a deep neural net that estimates light parameters for individual light objects. This method is the closest to our work. They used the Laval Indoor HDR dataset~\cite{Gardner17} which contains about 2100 \ac{HDR} maps to train the network. These parameters for the training data were determined by fitting ellipses on the \ac{HDR} intensity maps. The brightest area in the map was detected and the ellipse then was fitted by region growing. This area was masked and the process was repeated to determine a number of light sources. The parameters of the light source were defined by the size of the ellipse, average HDR intensity in the ellipse area and average HDR color value. Furthermore, a predicted depth map was used to determine the distance to the light source. We also use the Laval Indoor HDR dataset and with a DenseNet pretrained on ImageNet a similar network architecture. However, unlike Gardner et al. we estimate a light direction and therefore do not need to rely on predicted depth maps for the dataset.

\textbf{Shadow calculation} is a very broad and relatively old field of research in computer graphics. It ranges from simple methods for computing hard shadows, such as projection shadows~\cite{Blinn1988}, shadow mapping~\cite{Williams1978} and shadow volumes~\cite{Crow77} to more advanced methods for computing soft shadows like image-based soft shadows~\cite{Agrawala2000}, geometry-based soft shadows~\cite{Parker1998} and volumetric shadows~\cite{Kajiya1984}.

In contrast to previous work, we present a new approach in which we encode pre-computed shadow textures for an object in the weights of a neural network. This has the advantage that realistic soft shadows can be displayed in real-time on mobile devices, since the network can be queried very quickly. The idea of encoding images or textures in neural networks is not new. Stanley~\cite{Stanley2007} encoded image information in Compositional Pattern Producing Network (CPPN) inspired by encoding in natural DNA. Rainer et al.~\cite{Rainer2019, Rainer2020} used neural networks to compress the \ac{BTF}. Mildenhall et al.~\cite{NeRF2021} trained an \ac{MLP} to generate novel views from unknown perspectives of complex scenes. They used a mapping for the input coordinates to create a higher dimensional input space that allowed more high frequency variations in their output. This strategy was inspired by the positional encoding in the Transformer architecture~\cite{Transformer2017} and is also used by our method.

\section{Light estimation}
\label{sec:light}
To estimate the existing light situation in a scene using a single RGB image, we characterize the light situation by a set of parameters
\begin{equation}
	\left( \bm{d}, \bm{c}, \bm{a}, o \right).
\end{equation}
Here $\bm{d} \in \mathbb{R}^3$ is a unit vector which determines the light direction, $\bm{c}\in \mathbb{R}^3$ is the light color defined by RGB values with normalized components in $[\text{0},\text{1}]$ and $\bm{a}$ is an RGB vector corresponding to the ambient lighting of the scene. The parameter $o$ is a scalar value and measure for the opacity of the shadow texture described in \Cref{sec:shadows}. A value $o =$ 1 corresponds to an alpha value of 100\% and a value of $o =$ 0 corresponds to an alpha value of 0\%, i.e. invisible shadows. We train a \ac{CNN} to predict these parameters from a single RGB image with a resolution of 256x192 pixels.

\subsection{Input Data}
\label{subsec:light-data}
\begin{figure}
	\centering
	\setlength{\belowcaptionskip}{0.25\baselineskip}
	\begin{subfigure}{0.95\linewidth}
		\includegraphics[width=.95\linewidth]{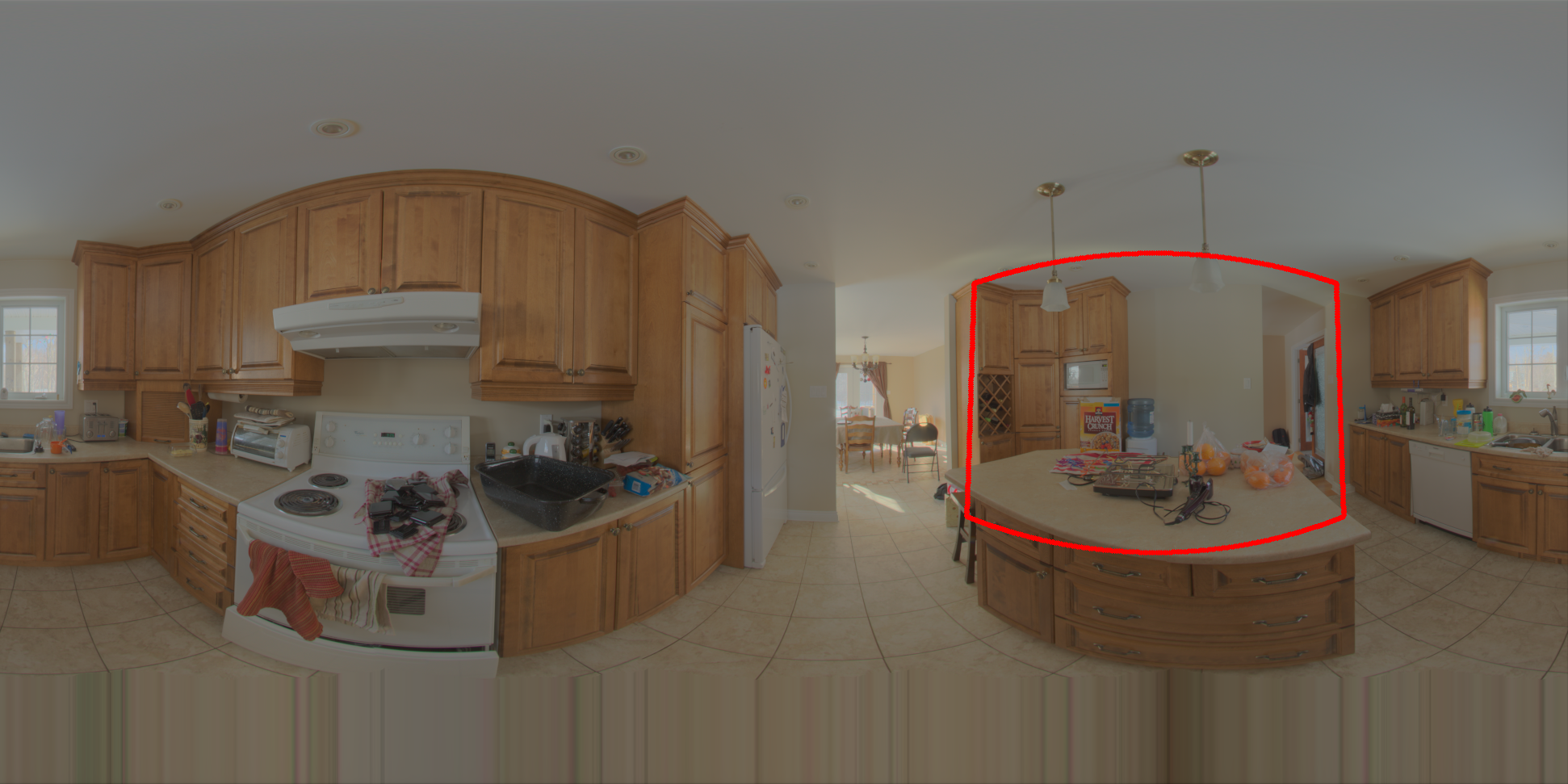}
		\caption{Original panorama}
		\label{fg:panoramas-a}
	\end{subfigure}
	\centering
	\begin{subfigure}{0.36\linewidth}
		\includegraphics[width=.95\linewidth]{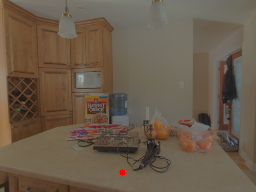}
		\caption{Cropped image}
		\label{fg:panoramas-b}
	\end{subfigure}
	\begin{subfigure}{0.54\linewidth}
		\includegraphics[width=.95\linewidth]{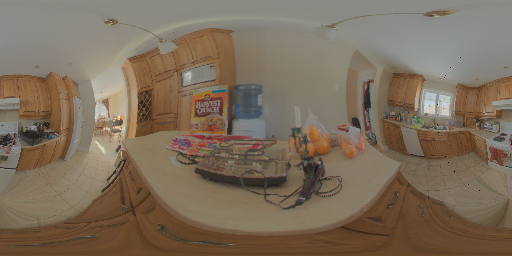}
		\caption{Warped panorama}
		\label{fg:panoramas-c}
	\end{subfigure}
	\caption{For a given panorama (a), the image information from inside the red frame is used to create the rectified cropped image (b). A warped panorama (c) is projected around the insertion point (red point in (b)).}
	\label{fg:panoramas}
\end{figure}
For training the network, a large number of images is needed for which the exact light situation of the whole scene is known. 360$^\circ$ \ac{HDR} panoramas are particularly suitable for this, since one can crop a limited field-of-view image from them to obtain input images (see \cref{fg:panoramas-a} \& \cref{fg:panoramas-b}), while still being able to recover the entire lighting situation of the scene. We use the Laval Indoor HDR dataset~\cite{Gardner17} which contains about 2100 \ac{HDR} panoramas, taken at different indoor scenes.

Like Gardner et. al~\cite{Gardner17}, we extract 8 different limited field-of-view crops per panorama at random polar angles $\theta$ between 60$^{\circ}$ and 120$^{\circ}$ and azimuth angles $\phi$ between 0$^{\circ}$ and 360$^{\circ}$. We use a field-of-view (FOV) of 85$^{\circ}$ to approximate the viewing angle of non-wide-angle cameras in modern smartphones. We perform a rectilinear projection (see red frame in \cref{fg:panoramas-a})) to back-project the distortion in the panoramas. The 360$^{\circ}$ \ac{HDR} panorama describes the light situation of the whole scene at the point where the camera was placed for the panorama. However, this does not correspond to the exact lighting situation around the cropped image. For finding out the exact light situation at that point, one would have to shoot a new 360$^{\circ}$ \ac{HDR} panorama at the virtual camera location of the cropped image. To estimate the light situation at this location we rotate the original panorama so that the cropped area is exactly in the center and then apply the same warping operator as described in ~\cite{Gardner17}. The resulting new panorama (see \cref{fg:panoramas-c}) is an approximation of the panorama around the virtual camera location of the cropped area.

We use each of the warped panoramas to extract the light parameters for the corresponding cropped input image. We first determine the pixel intensity $I_{ij}$ by adding the individual RGB channels with weights that correspond to the natural perception of the individual colors, i.e.
\begin{equation}
	I_{ij} = 0.0722 \cdot R_{ij} + 0.7152 \cdot G_{ij} + 0.2126 \cdot B_{ij},
\end{equation}
where $i$ is the pixel's column and $j$ its row.

Then we mask the areas where the intensity is greater than 5\% of the maximum intensity $I_{\text{max}}$ as highlights. It should be noted that this is only applicable when working with \ac{HDR} data. To determine the average light direction from the highlight area, we introduce two weights. First, the light direction of each pixel is weighted by its intensity. Second, the light direction of each pixel is weighted by the area that this pixel occupies on the unit sphere:
\begin{equation}
	\omega_{ij} = \frac{2\pi^2}{w \cdot h} \sin \left( \frac{j + 0.5}{h} \pi \right)
\end{equation}
where $j$ is the pixel's row and $w$, $h$ are the width and height of the panorama. This is necessary because, for example, an area near the poles occupies significantly more pixels on the panorama than an area with the same size at the equator. The resulting average light direction is the parameter $\bm{d}$. To determine the light color $\bm{c}$, the same weights are applied to the individual RGB values of the highlight area in a tone-mapped version of the panorama to obtain a mean highlight color. The ambient color $\bm{a}$ can be determined from the remaining pixel values of the tone-mapped panorama by using the same procedure. We determine the value for the opacity parameter $o$ from the quotient of the summed weighted intensities for the highlight areas $I^{\text{tot}}_{l}$ and analog for the remaining areas $I^{\text{tot}}_{a}$:
\begin{equation}
	o = 1 - \tanh\left(\frac{I^{\text{tot}}_{a}}{0.05 \cdot I^{\text{tot}}_{l} }\right).
\end{equation}
The less the intensities from the highlight areas differ from those of the ambient area, the lower the opacity of the shadow textures.

\subsection{Network Architecture}
\begin{figure}
	\centering
	\includegraphics[width=1.\linewidth]{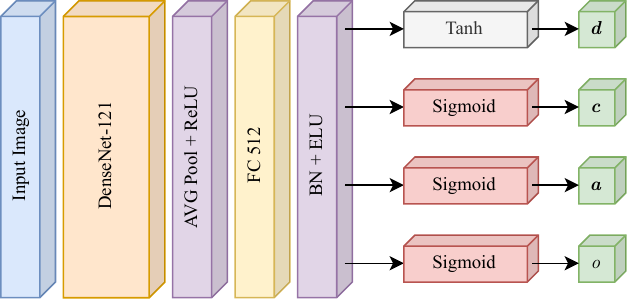}
	\caption{Proposed light estimation network architecture.}
	\label{fg:light_net}
\end{figure}
As mentioned before we use a \ac{CNN} to estimate the parameters from the input RGB image. Since the dataset is too small to train a network from scratch, we use a DenseNet-121~\cite{DenseNet2017}, pretrained on ImageNet~\cite{ImageNet2009} as an encoder. The block configuration is (6, 12, 24, 16) with a growth rate of 32, a compression of 0.5 and a batch norm size of 4. Furthermore, 64 initial features, ReLU activations and 2D average pooling with a pool size of 4 are used. The classifier of the DenseNet is removed, so the network produces a latent vector with size 512. This is forwarded to a fully connected (FC) 512 layer with batch norm and ELU activation. For each of the four parameters there is a separate FC layer as network head. The heads for the parameters $\bm{c}$, $\bm{a}$, $o$ are each normalized using a sigmoid function so that they lie between $(\text{0},\text{1})$. For the parameter $\bm{d}$ we use a tanh activation function and normalize the entire vector to unit length. The complete architecture is visualized in \Cref{fg:light_net}.

\subsection{Training \& Implementation}
\label{sec:lighttrain}
During training, we directly compare the estimated parameters with the ground truth parameters. Thereby individual losses for each head are calculated as mean squared error. The total loss function is the weighted sum of the individual losses, i.e.
\begin{align}
	\begin{split}
		\mathcal{L} \;=\; 
		&  \omega_d l_2(\bm{d}^{\text{est}}, \bm{d}^{\text{gt}}) 
		+ \omega_c l_2(\bm{c}^{\text{est}}, \bm{c}^{\text{gt}}) \\
		& + \omega_a l_2(\bm{a}^{\text{est}}, \bm{a}^{\text{gt}}) 
		+ \omega_o l_2(o^{\text{est}}, o^{\text{gt}}).
	\end{split}
\end{align}

We weight the individual losses differently with the weights $\omega_d =$ 5, $\omega_c =$ 2, $\omega_a =$ 2 and $\omega_o =$ 1. Since a correct estimation of the direction is of utmost importance for us, $\omega_d$ gets the highest value. 

We train the network for a total of 60 epochs using an Adam optimizer with $\beta_1 =$ 0.9 and $\beta_2 =$ 0.999. The learning rate $l_r =$ 0.001 is halved every 15 epochs. We use a batch size of 128 samples and a random 85/15 split of the dataset for training and validation. Scenes unknown to the network were used for testing. Typically, training takes about 2 hours on two Nvidia RTX A6000 GPUs. In total, our network consists of 7.7M parameters. The interference time on the iPhone 11 Pro GPU is 62ms and on the \ac{ANE} 9ms.

\subsection{Evaluation}
\begin{figure*}
	\centering
	\begin{subfigure}{0.24\linewidth}
		\includegraphics[width=.95\linewidth]{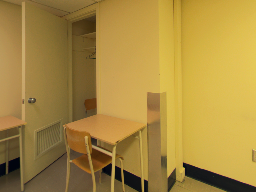}
	\end{subfigure}
	\hfill
	\begin{subfigure}{0.365\linewidth}
		\includegraphics[width=.95\linewidth]{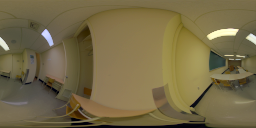}
	\end{subfigure}
	\hfill
	\begin{subfigure}{0.185\linewidth}
		\includegraphics[width=.95\linewidth]{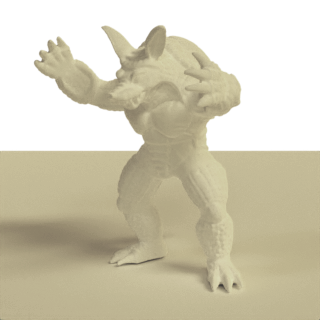}
	\end{subfigure}
	\hfill
	\begin{subfigure}{0.185\linewidth}
		\includegraphics[width=.95\linewidth]{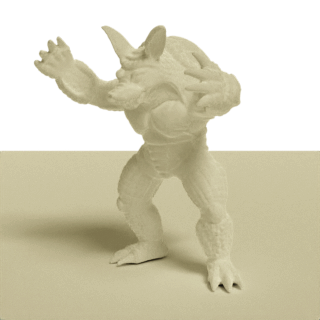}
	\end{subfigure}
	\par\medskip
	\centering
	\begin{subfigure}{0.24\linewidth}
		\includegraphics[width=.95\linewidth]{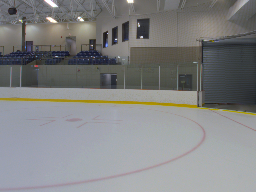}
	\end{subfigure}
	\hfill
	\begin{subfigure}{0.365\linewidth}
		\includegraphics[width=.95\linewidth]{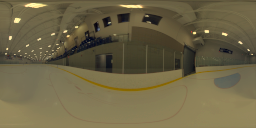}
	\end{subfigure}
	\hfill
	\begin{subfigure}{0.185\linewidth}
		\includegraphics[width=.95\linewidth]{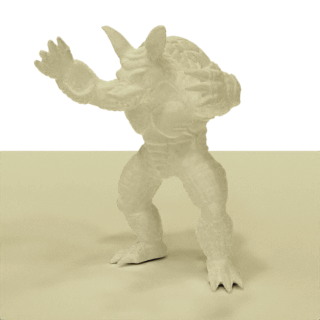}
	\end{subfigure}
	\hfill
	\begin{subfigure}{0.185\linewidth}
		\includegraphics[width=.95\linewidth]{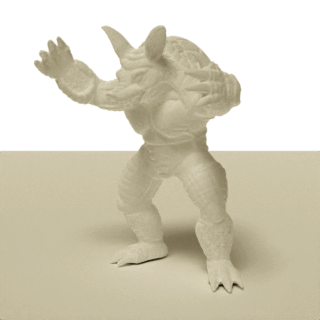}
	\end{subfigure}
	\par\medskip
	\centering
	\setlength{\belowcaptionskip}{0.25\baselineskip}
	\begin{subfigure}{0.24\linewidth}
		\includegraphics[width=.95\linewidth]{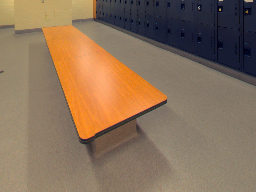}
		\caption{Input image}
		\label{fg:eval_light_a}
	\end{subfigure}
	\hfill
	\begin{subfigure}{0.365\linewidth}
		\includegraphics[width=.95\linewidth]{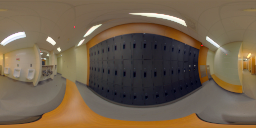}
		\caption{GT HDR panorama}
		\label{fg:eval_light_b}
	\end{subfigure}
	\hfill
	\begin{subfigure}{0.185\linewidth}
		\includegraphics[width=.95\linewidth]{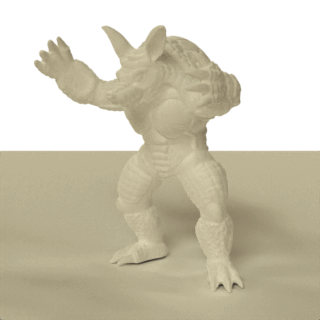}
		\caption{GT result}
		\label{fg:eval_light_c}
	\end{subfigure}
	\hfill
	\begin{subfigure}{0.185\linewidth}
		\includegraphics[width=.95\linewidth]{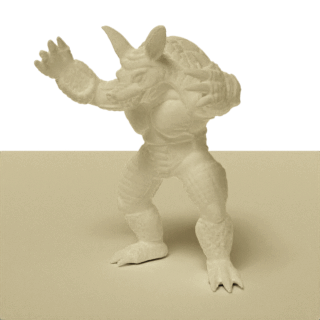}
		\caption{Our result}
		\label{fg:eval_light_d}
	\end{subfigure}
	\caption{Exemplified representation of our evaluation. (a) shows the input image, (b) the corresponding GT HDR panorama, (c) the GT image of the Armadillo rendered with IBL techniques, and (d) the image of the same Armadillo rendered using the light parameters from our light estimation.}
	\label{fg:eval_light}
\end{figure*}
\begin{table}
	\centering
	\begin{tabularx}{0.8\linewidth}{l|c|c}
		\toprule
		\textbf{Metric} & \textbf{Gardner19}(1) & \textbf{Ours} \\
		\hline
		RMSE & 0.1114 & \textbf{0.1101} \\
		si-RMSE & 0.1518 & \textbf{0.1501} \\
		RMLE & 0.07007 & \textbf{0.06928}\\
		Angular Error & 3.556$^\circ$ & \textbf{3.542$^\circ$}\\
		\bottomrule
	\end{tabularx}
	\caption{Comparision by different widely used metrics of our method with the state of the art parametric indoor light estimation by Gardner et al.~\cite{Gardner19} with one light source. Best results in bold.}
	\label{tb:eval_light}
\end{table}
Comparing the results of two light estimation approaches is challenging. Since qualitative evaluation always contains personal bias, we rely on a purely quantitative measure for this evaluation. We don't compare our method with approaches, that do not estimate a parametric light direction but spatially varying light coefficients like spherical harmonics~\cite{Cheng18, Garon19} or complete environment maps~\cite{Gardner17, Song19, Somanath21}, because we especially need the light direction for the shadow calculation (\cref{sec:shadows}). We therefore compare our light estimation approach only with the work of Gardner et al.~\cite{Gardner19} when using one main light source. We neglect our opacity parameter $o$ at this point, since its use is mainly for the shadow textures presented in \Cref{sec:shadows} and will be evaluated in the overall pipeline evaluation in \Cref{sec:eval}.

We use a simple scene with an armadillo and a plane as a shadow catcher (see \cref{fg:eval_light_c}). For a given input image (see \cref{fg:eval_light_a}), we render a GT image (see \cref{fg:eval_light_c}) with the corresponding warped GT environment map (\cref{fg:eval_light_b}), as described in \Cref{subsec:light-data}, with IBL techniques. We then estimate light parameters with the respective light estimation. The same scene is rendered again with a parametric light source and ambient color (see \cref{fg:eval_light_d}). 

To compare renderings of the two predictions with the GT image we use 4 different metrics. On the one hand RMSE as well as the scale-invariant si-RMSE and RMLE and on the other hand a per pixel RGB angular error~\cite{AngularError2004}. The standard RMSE is a good measure for the error in the relation between ambient and light intensity. The two scale-invariant measures filter out differences in the scales of the two images and are therefore good measures for errors in light position due to difference in shadows. The RGB angular error, on the other hand, comes from whitebalance research and is a good measure to evaluate the color predicition of the light source and the ambient color.

In total, we evaluated 977 images from a test set unknown to the network. We used Blender~\cite{Blender} for all renderings. \Cref{tb:eval_light} shows the results of our evaluation. It can be seen that our method performs 1-1.2\% better than the previous state-of-the-art method in all metrics when considering only a single light source.

\section{Shadows}
\label{sec:shadows}

\begin{figure}
	\centering
	\setlength{\belowcaptionskip}{0.25\baselineskip}
	\begin{subfigure}{0.55\linewidth}
		\includegraphics[width=1.0\linewidth]{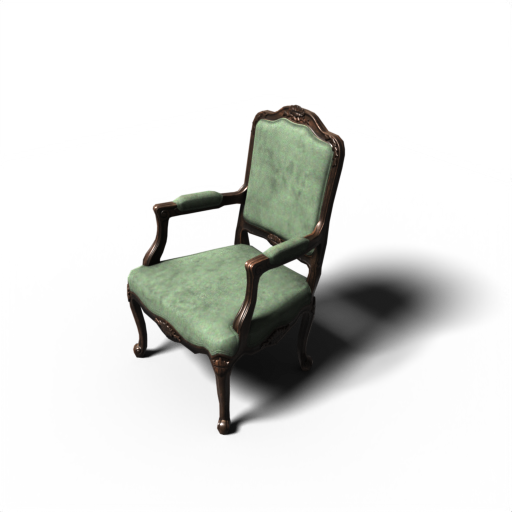}
		\caption{Object}
		\label{fg:planar-a}
	\end{subfigure}
	\hfill
	\begin{subfigure}{0.39\linewidth}
		\includegraphics[width=1.0\linewidth]{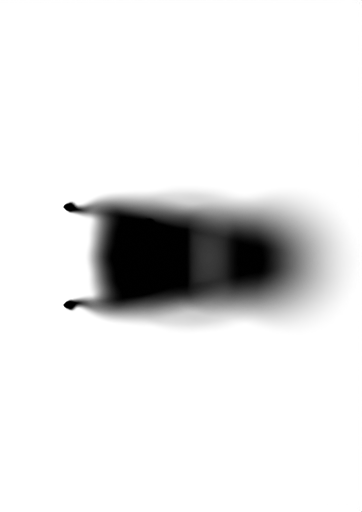}
		\caption{Shadow texture}
		\label{fg:planar-b}
	\end{subfigure}

	\caption{A chair lit by a front light (a) with the corresponding shadow texture (b).}
	\label{fg:planar}
\end{figure}

We aim to generate a planar shadow texture (see \cref{fg:planar-b}), i.e. a 2D grayscale image, depending on the light direction defined by a unit vector $\bm{d} \in \mathbb{R}^3$ for a specific object (see \cref{fg:planar-a}). Our experiments showed that the use of cartesian coordinates leads to a more stable training than spherical coordinates since the network seems to have problems with the discontinuity between $\phi =$ 2$\pi$ and $\phi =$ 0. This results in a \emph{shadow function} 
\begin{equation}
	f : \mathbb{R}^5 \longrightarrow \mathbb{R},\quad  f(i,j,\bm{d}) \longrightarrow v,
	\label{eq:pix_func}
\end{equation}
that maps pixel position $(i,j)$ together with a light direction $\mathbf{d}$
to a grayscale value $v$. We use a \ac{MLP}s are a universal function approximator~\cite{Hornik1989}, to represent the desired shadow function. 

\subsection{Input Data}
\begin{figure}
	\centering
	\includegraphics[width=1.0\linewidth]{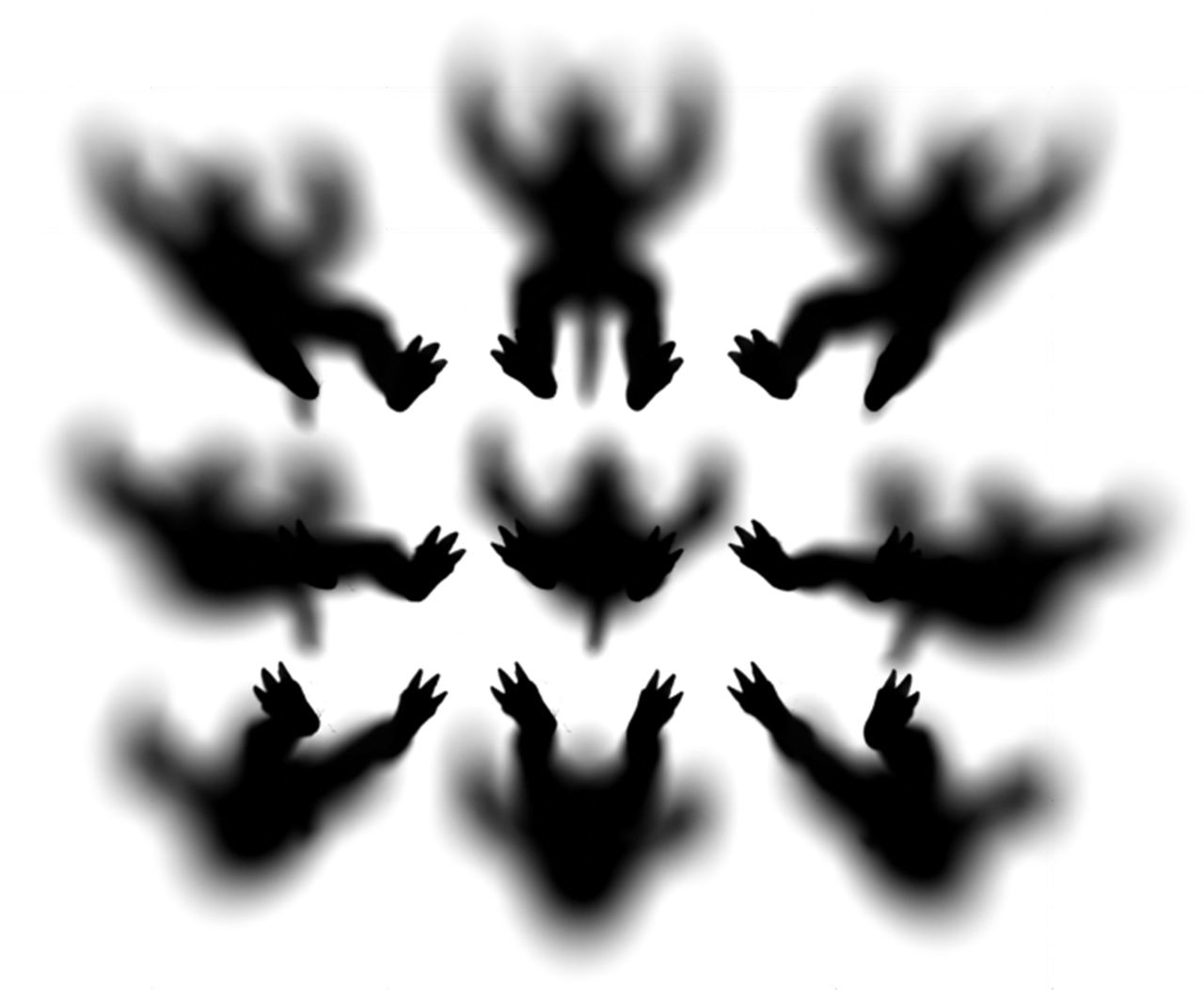}
	\caption{Shadow textures of the Armadillo (see \cref{fg:eval_light_c}) from different light directions $\bm{d}$.}
	\label{fg:textures}
\end{figure}

We train one specific network for each individual model. As training data, we use shadow textures for a variety of different light directions. These textures are created with a simple scene setup and the Cycles render engine in Blender~\cite{Blender}. The scene consists of a quadratic plane that acts as a shadow catcher. The plane is dimensioned so that its side length is three times as long as the largest side of the bounding box that contains the object to be trained. The object is centered on the plane and is assigned a material that is invisible to the render engine but allows shadow casting. An orthographic camera from the top view captures the textures. A directional light (sun in blender) with an opening angle of 20$^{\circ}$ is used as the light source. This type of light is defined by one direction and still produces soft shadows. It's therefore well suited as an approximation for indoor shadows. This light source is set to different light directions for the individual training samples. We use uniformly distributed spherical angles $\theta$, $\phi$. Where $\theta$ takes values from 0$^{\circ}$ to 45$^{\circ}$ with an increment of 4.5$^{\circ}$ and $\phi$ takes values from $0^{\circ}$ to 360$^{\circ}$ with an increment of 12$^{\circ}$. This results in a total of 301 texture samples. For each sample, we use a resolution of 256x256 pixels. \Cref{fg:textures} shows an example of shadow textures for different light directions for the Armadillo (see \cref{fg:eval_light_c}).
\begin{figure*}
	\centering
	\setlength{\belowcaptionskip}{0.25\baselineskip}
	\begin{subfigure}{0.49\linewidth}
		\includegraphics[width=.95\linewidth]{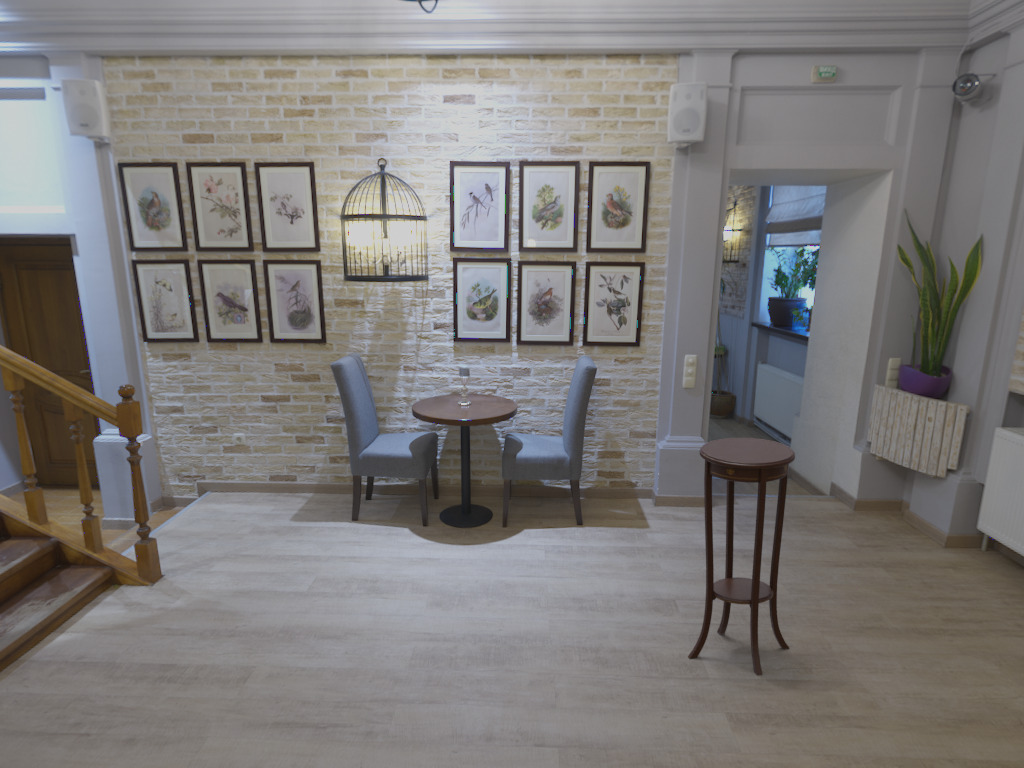}
		\caption{Ground truth}
		\label{fg:overall-eval-a}
	\end{subfigure}
	\hfill
	\begin{subfigure}{0.49\linewidth}
		\includegraphics[width=.95\linewidth]{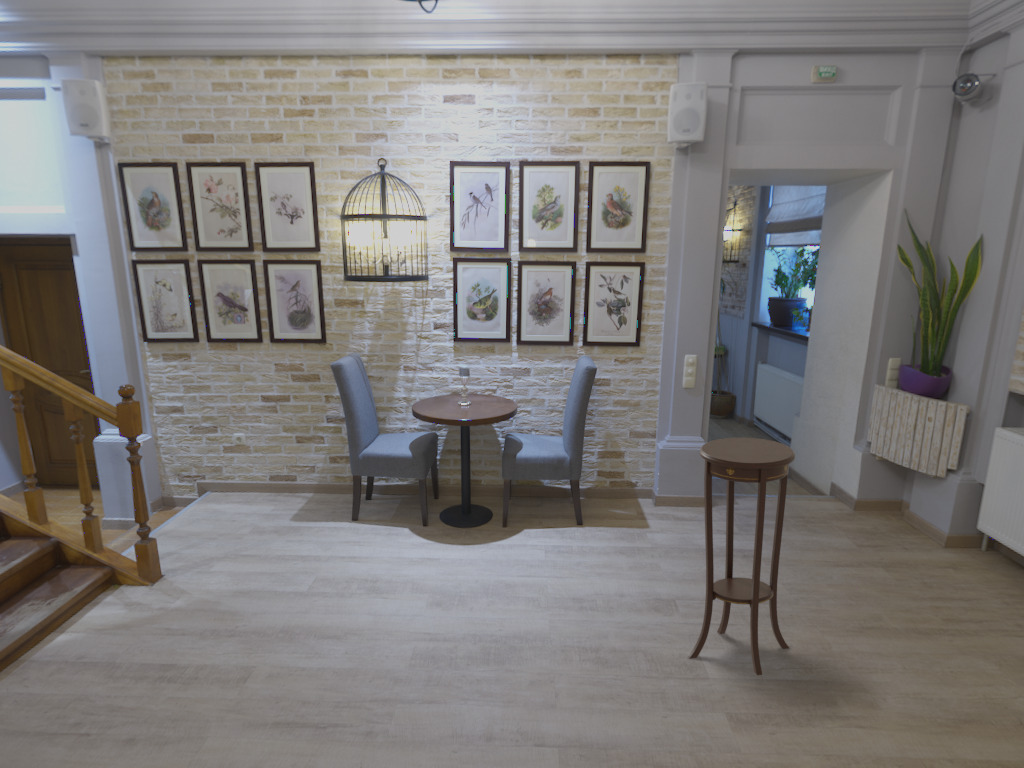}
		\caption{Ours}
		\label{fg:overall-eval-b}
	\end{subfigure}
	\caption{An example of our qualitative evaluation. Left: Coffee table rendered with the ground truth HDR panorama around the insertion point. Right: Coffee table rendered with a directional light and ambient color from our light estimation and shadow texture from our neural soft shadows.}
	\label{fg:overall-eval}
\end{figure*}

\subsection{Network Architecture}
\begin{figure}
	\centering
	\includegraphics[width=1.\linewidth]{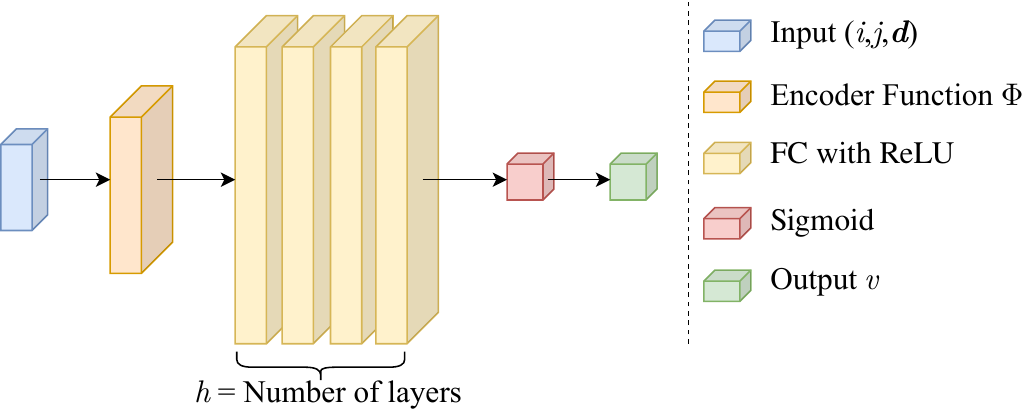}
	\caption{Proposed shadow network architecture.}
	\label{fg:shadow_net}
\end{figure}
As mentioned before (see \cref{eq:pix_func}), all information about the shadows is mapped by pixel-wise functions from 5D space to 1D grayscale information. Since neural networks tend to learn a low-frequency bias, we assist them in learning high-frequency details by mapping the 5D input to a higher dimensional space, as shown by Rahaman et al.~\cite{Rahaman2018}. This technique is also used very successfully with NeRFs~\cite{NeRF2021}. Similar to Vaswani et al.~\cite{Transformer2017} with Transformers, we use an encoder function $\Phi$ to map each of the five input dimensions $x \in \mathbb{R}$ to a higher dimensional sequence of alternating sine and cosine functions:
\begin{align}
	\begin{split}
		\Phi(x) =  & \left( \sin(2^0 \pi x), \cos(2^0 \pi x), \dots, \right.  \\
		&\left. \dots, \sin(2^{L-1} \pi x), \cos(2^{L-1} \pi x) \right)
	\end{split}
\end{align}
where $L$ is a dimensionality parameter. The image space $(i,j)$ is normalized to values in [0,1]. For the image space   encoding is done with a dimensionality parameter $L =$ 10. The elements of the light direction vector $\bm{d}$ by definition take only values in [-1,1] and for their encoding we choose an $L =$ 4 analogous to the viewing direction vector in ~\cite{NeRF2021}. In total we map the $\mathbb{R}^5$ input space to higher dimensional space of $\mathbb{R}^{64}$. The input passes through $h$ hidden layers, each with a filter size $s$, and is activated with ReLUs after each hidden layer (see \cref{fg:shadow_net}). In our experiments we use a filter size $s$ of 128 to 256 and a number of hidden layers $h$ from 1 to 4. The output value $v$ of the network is normalized with a sigmoid function between 0 and 1.

\subsection{Training}
During training, for each shadow texture sample $k$ with fixed light direction $\bm{d}$, we take a number of $N$ random continuous pixel locations $(i,j) \in [\text{0},\text{1}]$. Here, the ground truth grayscale value $v_{i,j}^{\text{gt}}$ at the continuous location $(i,j)$ is obtained by bilinear interpolating from the known values at the discrete surrounding known pixel values. It should be mentioned that it is also possible to train the network without interpolation only on random known discrete pixel values. This speeds up the training by a factor of 5 since filtering is a bottleneck. On the other hand, it reduces the quality of the network, and the ability to predict different resolutions with the network is lost. As loss function $\mathcal{L}$ we take the mean squared error loss $l_2$ between estimated pixel value $v^{\text{est}}$ and interpolated pixel value $v^{\text{gt}}$:
\begin{equation}
	\mathcal{L} = l_2(v^{\text{est}}, v^{\text{gt}}).
\end{equation}

\subsection{Implementation}
One advantage of our method is that the resulting network is very small and thus not only requires little memory, but a forward pass also has a low interference time. The forward passes for all 65536 pixels of a 256x256 texture need in total about 33ms on the GPU of the iPhone 11 Pro and 5ms on the \ac{ANE} 9ms. Assuming a filter size $s =$ 128 and a number of hidden layers $h =$ 3 the network has just 58k parameters. The data set with its 301 grayscale images with a resolution of 256x256 is small enough to be loaded completely into the memory even with simple consumer GPUs. We train our network for a total of 10000 epochs and need about 5 minutes (or just under a minute without bilinear filtering) on an Nvidia RTX A6000. As in \Cref{sec:lighttrain}, we again use an Adam optimizer with standard values of $\beta_1 =$ 0.9 and $\beta_2 =$ 0.999. We apply an exponential learning rate decay ($\gamma =$ 0.99977) to the initial learning rate $l_r =$ 0.001 so that it is reduced to one-tenth of the original value after 10000 epochs. Per texture sample we use $N =$ 256 pixel locations, which results in 77k network passes per epoch.

\subsection{Limitations}
Currently, our method is only suitable for creating a planar shadow texture for the plane it sits on. This is sufficient for of AR applications, where an object is placed in the middle of an empty room and is far enough away from walls to cast a shadow on them. Problems arise when a virtual object should cast shadows on another virtual object or on non-virtual objects in the scene. 
\section{Overall Pipeline Evaluation}
\label{sec:eval}
\begin{table}
	\centering
	\begin{tabularx}{0.66\linewidth}{l|c|c}
		\toprule
		& \textbf{GT} & \textbf{Ours} \\
		\hline
		Rating & 3.49 $\pm$ 0.38 & 3.26 $\pm$ 0.46 \\
		\hline
		Votes & 0.544\% & 0.456\% \\
		\bottomrule
	\end{tabularx}
	\caption{Results of the qualitative evaluation (20 images, 50 participants). Rating describes how realistically an objects fits into the scene considering only lighting and shadows on a scale from 1 (very unrealistic) to 5 (very realistic). Votes denotes the percentage of which image was prefered in terms of realistic look (50\% = perfect confusion).}
	\label{tb:eval_overall}
\end{table}
We determine the overall quality of our entire pipeline with a qualitative evaluation. For this we use new HDR panoramas that are not from the Laval Indoor HDR dataset and have not yet been seen by the network. For each panorama we choose a cropped rectified image where a virtual object should be inserted. We use the light estimation from \Cref{sec:light} to determine the light direction, light color, ambient color and the opacity value for the shadows. We then use the light direction to determine the shadow texture using our method from  \Cref{sec:shadows}. We insert the object into the image and render it using only a directional light and ambient lighting. We also add the neural shadow texture with the estimated opacity (see \cref{fg:overall-eval-a}). In comparison, we determine the warped panorama (see \cref{subsec:light-data}) at the insertion point and render the same object with ray traced \ac{IBL} and a plane as shadow catcher (see \cref{fg:overall-eval-b}).

A total of 20 images (see supplementary material) were created for qualitave evaluation. We showed these images to 50 participants. On the one hand, the participants were asked to assess how realistically an object fits into the existing scene in terms of its lighting and shadows. For the rating, we use the Likert scale with values from 1 (very unrealistic) to 5 (very realistic). Explicitly the participants were told not to consider syle, proportions, object selection and context. On the other hand, the participants were shown both pictures (see \cref{fg:overall-eval}) next to each other and they were asked to decide which of the two pictures they thought was more realistic looking in terms of lighting and shadows. \Cref{tb:eval_overall} shows the results of our survey. It turns out that the participants as a whole give the ground truth visualizations only a slightly higher quality rating than our visualizations. This is also confirmed by the fact that quite a few participants prefer our visualization to the ground truth in a direct comparison.

\begin{figure}
	\centering
	\setlength{\belowcaptionskip}{0.25\baselineskip}
	\begin{subfigure}{0.95\linewidth}
		\includegraphics[width=.95\linewidth]{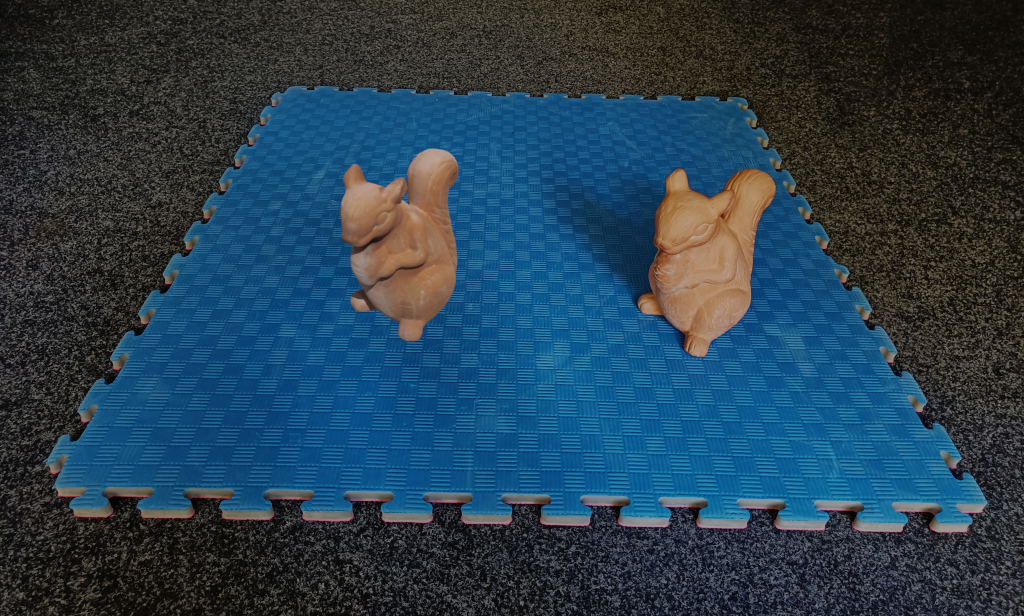}
		\caption{Without shadow cast}
		\label{fg:compare-a}
	\end{subfigure}
	\centering
	\begin{subfigure}{0.95\linewidth}
		\includegraphics[width=.95\linewidth]{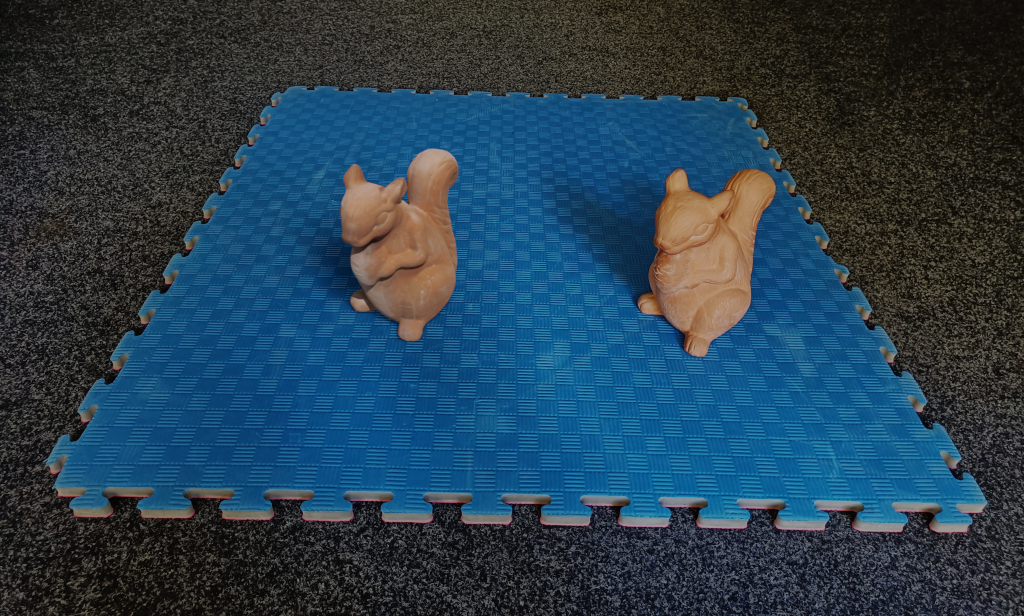}
		\caption{With neural shadow texture}
		\label{fg:compare-b}
	\end{subfigure}
	\caption{Comparison between a real clay squirrel (right) and the virtual object (left) rendered with the light parameter of the light estimation from \Cref{sec:light}. (a) shows the object without shadow cast and (b) with the neural shadow texture from our method in \Cref{sec:shadows}.}
	\label{fg:compare}
\end{figure}

Furthermore, in \Cref{fg:compare} we compare a real object with a rendered virtual version. For this we place a real clay squirrel in the room and leave space for the virtual version. The photo was taken with an ordinary smartphone and the light estimation from \Cref{sec:light} was used to determine the light direction, light color, ambient color and the opacity value of the shadows. The virtual squirrel was inserted on the left and rendered with the light parameters. \Cref{fg:compare-a} shows the virtual squirrel without shadow cast. \Cref{fg:compare-b} shows the squirrel with the neural soft shadow texture generated with our method from \Cref{sec:shadows}. It is easy to see that without shadows the object looks out of place in the scene. The subtle soft shadow of our method, on the other hand, conveys immersion.

\balance

\section{Conclusion}
\label{sec:conclusion}
We presented a complete pipeline for realistic embedding of virtual objects into indoor scenes. Our light estimation determines a parametric description of the light situation from an RGB image as input. Our neural soft shadow method generates realistic soft shadows as textures that allow to embed virtual objects in previously unknown levels of realismn in real-time into AR scenes. Of course, our method is not suitable for reproducing complex lighting situations exactly, but it is suitable for giving the viewer a convincing sense of immersion. This is supported by our user test where approximately the same number of subjects preferred our method over ground truth visualization. Our entire pipeline is real-time capable on current mobile devices. 

In particular, our fundamental work in the area of neural soft shadows opens up a wide range of possibilities for future research. At the moment we are working on how to effectively transfer our method to the shadow cast on walls. In this case, the distance to the wall adds another degree of freedom to the problem. It would be interesting to incorporate more complex light sources, such as area lights, with further parameters like light size in neural shadows. It is also exciting to see if multiple light sources can be represented as neural soft shadows. Furthermore, we could imagine that complex shadows of semi-transparent objects could be another future application of our method.

\section*{ACKNOWLEDGMENTS}
\censor{This project (HA project no. 1102/21-104) is financed with funds of LOEWE - Landes-Offensive zur Entwicklung Wissenschaftlich-\"okonomischer Exzellenz, F\"orderlinie 3: KMU-Verbundvorhaben (State Offensive for the Development of Scientific and Economic Excellence).}
\pagebreak
\balance

\bibliographystyle{plain}

\footnotesize
\bibliography{references}

\begin{thebibliography}{10}

\bibitem{Agrawala2000}
Maneesh Agrawala, Ravi Ramamoorthi, Alan Heirich, and Laurent Moll.
\newblock Efficient image-based methods for rendering soft shadows.
\newblock In {\em Proceedings of the 27th Annual Conference on Computer
  Graphics and Interactive Techniques}, SIGGRAPH '00, pages 375--384, USA,
  2000. ACM Press/Addison-Wesley Publishing Co.

\bibitem{Balci17}
Hasan Balc\i\ and U\u{g}ur G\"ud\"ukbay.
\newblock Sun position estimation and tracking for virtual object placement in
  time-lapse videos.
\newblock {\em Signal, Image and Video Processing}, 11, 07 2017.

\bibitem{Barron2013}
Jonathan~T. Barron and Jitendra Malik.
\newblock Shape, illumination, and reflectance from shading.
\newblock {\em IEEE Transactions on Pattern Analysis and Machine Intelligence},
  37:1670--1687, 2013.

\bibitem{Blinn1988}
James~F. Blinn.
\newblock Me and my (fake) shadow.
\newblock {\em IEEE Computer Graphics and Applications}, 8:82--86, 1988.

\bibitem{Cheng18}
Dachuan Cheng, Jian Shi, Yanyun Chen, Xiaoming Deng, and Xiaopeng Zhang.
\newblock Learning scene illumination by pairwise photos from rear and front
  mobile cameras.
\newblock {\em Computer Graphics Forum}, 37:213--221, 10 2018.

\bibitem{Blender}
Blender~Online Community.
\newblock {\em Blender - a 3D modelling and rendering package}.
\newblock Blender Foundation, Stichting Blender Foundation, Amsterdam, 2022.

\bibitem{Crow77}
Franklin~C. Crow.
\newblock Shadow algorithms for computer graphics.
\newblock In {\em Proceedings of the 4th Annual Conference on Computer Graphics
  and Interactive Techniques}, SIGGRAPH '77, pages 242--248, New York, NY, USA,
  1977. Association for Computing Machinery.

\bibitem{Debevec98}
Paul Debevec.
\newblock Rendering synthetic objects into real scenes: Bridging traditional
  and image-based graphics with global illumination and high dynamic range
  photography.
\newblock In {\em Proceedings of the 25th Annual Conference on Computer
  Graphics and Interactive Techniques}, SIGGRAPH '98, pages 189--198, New York,
  NY, USA, 1998. Association for Computing Machinery.

\bibitem{Debevec97}
Paul~E. Debevec and Jitendra Malik.
\newblock Recovering high dynamic range radiance maps from photographs.
\newblock In {\em Proceedings of the 24th Annual Conference on Computer
  Graphics and Interactive Techniques}, SIGGRAPH '97, pages 369--378, USA,
  1997. ACM Press/Addison-Wesley Publishing Co.

\bibitem{ImageNet2009}
Jia Deng, Wei Dong, Richard Socher, Li-Jia Li, Kai Li, and Li~Fei-Fei.
\newblock Imagenet: A large-scale hierarchical image database.
\newblock In {\em 2009 IEEE Conference on Computer Vision and Pattern
  Recognition}, pages 248--255, 2009.

\bibitem{Gardner19}
Marc-Andr\'{e} Gardner, Yannick Hold-Geoffroy, Kalyan Sunkavalli, Christian
  Gagn\'{e}, and Jean-Fran\c{c}ois Lalonde.
\newblock Deep parametric indoor lighting estimation.
\newblock In {\em 2019 IEEE/CVF International Conference on Computer Vision
  (ICCV)}, pages 7174--7182, 10 2019.

\bibitem{Gardner17}
Marc-Andr\'{e} Gardner, Kalyan Sunkavalli, Ersin Yumer, Xiaohui Shen, Emiliano
  Gambaretto, Christian Gagn\'{e}, and Jean-Fran\c{c}ois Lalonde.
\newblock Learning to predict indoor illumination from a single image.
\newblock {\em ACM Trans. Graph.}, 36(6), nov 2017.

\bibitem{Garon19}
Mathieu Garon, Kalyan Sunkavalli, Sunil Hadap, Nathan Carr, and Jean-Francois
  Lalonde.
\newblock Fast spatially-varying indoor lighting estimation.
\newblock In {\em 2019 IEEE/CVF Conference on Computer Vision and Pattern
  Recognition (CVPR)}, pages 6901--6910, 06 2019.

\bibitem{Hold-Geoffroy17}
Yannick Hold-Geoffroy, Kalyan Sunkavalli, Sunil Hadap, Emiliano Gambaretto, and
  Jean-Fran\c{c}ois Lalonde.
\newblock Deep outdoor illumination estimation.
\newblock In {\em 2017 IEEE Conference on Computer Vision and Pattern
  Recognition (CVPR)}, pages 2373--2382, 07 2017.

\bibitem{AngularError2004}
S.D. Hordley and G.D. Finlayson.
\newblock Re-evaluating colour constancy algorithms.
\newblock In {\em Proceedings of the 17th International Conference on Pattern
  Recognition, 2004. ICPR 2004.}, volume~1, pages 76--79 Vol.1, 2004.

\bibitem{Hornik1989}
K.~Hornik, M.~Stinchcombe, and H.~White.
\newblock Multilayer feedforward networks are universal approximators.
\newblock {\em Neural Netw.}, 2(5):359--366, jul 1989.

\bibitem{DenseNet2017}
Gao Huang, Zhuang Liu, Laurens Van Der~Maaten, and Kilian~Q. Weinberger.
\newblock Densely connected convolutional networks.
\newblock In {\em 2017 IEEE Conference on Computer Vision and Pattern
  Recognition (CVPR)}, pages 2261--2269, 2017.

\bibitem{Kajiya1984}
James~T. Kajiya and Brian~P Von~Herzen.
\newblock Ray tracing volume densities.
\newblock In {\em Proceedings of the 11th Annual Conference on Computer
  Graphics and Interactive Techniques}, SIGGRAPH '84, pages 165--174, New York,
  NY, USA, 1984. Association for Computing Machinery.

\bibitem{Lombardi16}
Stephen Lombardi and Ko~Nishino.
\newblock Reflectance and illumination recovery in the wild.
\newblock {\em IEEE Trans. Pattern Anal. Mach. Intell.}, 38(1):129--141, jan
  2016.

\bibitem{Lopez-Moreno13}
Jorge Lopez-Moreno, Elena Garces, Sunil Hadap, Erik Reinhard, and Diego
  Guti\'{e}rrez.
\newblock Multiple light source estimation in a single image.
\newblock {\em Computer Graphics Forum}, 32, 12 2013.

\bibitem{NeRF2021}
Ben Mildenhall, Pratul~P. Srinivasan, Matthew Tancik, Jonathan~T. Barron, Ravi
  Ramamoorthi, and Ren Ng.
\newblock Nerf: Representing scenes as neural radiance fields for view
  synthesis.
\newblock {\em Commun. ACM}, 65(1):99--106, dec 2021.

\bibitem{Parker1998}
Steven Parker, Peter Shirley, and Brian Smits.
\newblock Single sample soft shadow.
\newblock {\em Tech. Rep. UUCS-98-019}, 10 1998.

\bibitem{Rahaman2018}
Nasim Rahaman, Aristide Baratin, Devansh Arpit, Felix Dr{\"a}xler, Min Lin,
  Fred~A. Hamprecht, Yoshua Bengio, and Aaron~C. Courville.
\newblock On the spectral bias of neural networks.
\newblock In {\em ICML}, 2018.

\bibitem{Rainer2020}
Gilles Rainer, Abhijeet Ghosh, Wenzel Jakob, and Tim Weyrich.
\newblock Unified neural encoding of btfs.
\newblock {\em Computer Graphics Forum}, 39:167--178, 05 2020.

\bibitem{Rainer2019}
Gilles Rainer, Wenzel Jakob, Abhijeet Ghosh, and Tim Weyrich.
\newblock Neural btf compression and interpolation.
\newblock {\em Computer Graphics Forum}, 38:235--244, 05 2019.

\bibitem{Ismar14}
Kai Rohmer, Wolfgang Büschel, Raimund Dachselt, and Thorsten Grosch.
\newblock Interactive near-field illumination for photorealistic augmented
  reality on mobile devices.
\newblock In {\em 2014 IEEE International Symposium on Mixed and Augmented
  Reality (ISMAR)}, pages 29--38, 2014.

\bibitem{Somanath21}
Gowri Somanath and Daniel Kurz.
\newblock Hdr environment map estimation for real-time augmented reality.
\newblock In {\em 2021 IEEE/CVF Conference on Computer Vision and Pattern
  Recognition (CVPR)}, pages 11293--11301, 06 2021.

\bibitem{Song19}
Shuran Song and Thomas Funkhouser.
\newblock Neural illumination: Lighting prediction for indoor environments.
\newblock In {\em 2019 IEEE/CVF Conference on Computer Vision and Pattern
  Recognition (CVPR)}, pages 6911--6919, 06 2019.

\bibitem{Stanley2007}
Kenneth~O. Stanley.
\newblock Compositional pattern producing networks: A novel abstraction of
  development.
\newblock {\em Genetic Programming and Evolvable Machines}, 8(2):131162, jun
  2007.

\bibitem{Transformer2017}
Ashish Vaswani, Noam Shazeer, Niki Parmar, Jakob Uszkoreit, Llion Jones,
  Aidan~N. Gomez, \L{}ukasz Kaiser, and Illia Polosukhin.
\newblock Attention is all you need.
\newblock In {\em Proceedings of the 31st International Conference on Neural
  Information Processing Systems}, NIPS'17, pages 6000--6010, Red Hook, NY,
  USA, 2017. Curran Associates Inc.

\bibitem{Williams1978}
Lance Williams.
\newblock Casting curved shadows on curved surfaces.
\newblock {\em SIGGRAPH Comput. Graph.}, 12(3):270--274, aug 1978.

\end{thebibliography}

\end{document}